\documentclass[lettersize,journal]{IEEEtran}
\usepackage{amsmath,amsfonts}
\usepackage{array}
\usepackage{xcolor}
\usepackage[caption=false,font=normalsize,labelfont=sf,textfont=sf]{subfig}
\usepackage{textcomp}
\usepackage{stfloats}
\usepackage{url}
\usepackage{verbatim}
\usepackage{graphicx}
\usepackage{cite}
\usepackage{colortbl}
\usepackage{dsfont}
\newcommand{\revise}[1]{\textcolor{black}{#1}}

\usepackage{times}  
\usepackage{helvet}  
\usepackage{courier}  
\usepackage[square,numbers]{natbib}  
\usepackage{caption} 
\usepackage{multirow}
\usepackage{adjustbox}

\usepackage[linesnumbered,ruled,vlined]{algorithm2e}
\usepackage{algpseudocode}
\usepackage{amsmath}

\begin{document}

\title{Exploring Robust Features for Improving Adversarial Robustness}

\author{Hong Wang, Yuefan Deng, Shinjae Yoo, Yuewei Lin*
\thanks{H. Wang, S. Yoo and Y. Lin are with Computational Science Initiative, Brookhaven National Laboratory, Upton, NY, USA.}
\thanks{Y. Deng is with the department of Applied Mathematics and Statistics, Stony Brook University, Stony Brook, NY, USA.}
\thanks{*Y. Lin is the corresponding author.}}

\markboth{Journal of \LaTeX\ Class Files,~Vol.~14, No.~8, August~2021}%
{Shell \MakeLowercase{\textit{et al.}}: A Sample Article Using IEEEtran.cls for IEEE Journals}


\maketitle

\begin{abstract}
While deep neural networks (DNNs) have revolutionized many fields, their fragility to carefully designed adversarial attacks impedes the usage of DNNs in safety-critical applications. In this paper, we strive to explore the robust features which are not affected by the adversarial perturbations, i.e., invariant to the clean image and its adversarial examples, to improve the model's adversarial robustness. Specifically, we propose a feature disentanglement model to segregate the robust features from non-robust features and domain specific features. The extensive experiments on \revise{four} widely used datasets with different attacks demonstrate that robust features obtained from our model improve the model's adversarial robustness compared to the state-of-the-art approaches.
Moreover, the trained domain discriminator is able to identify the domain specific features from the clean images and adversarial examples almost perfectly. This enables adversarial example detection without incurring additional computational costs. With that, we can also specify different classifiers for clean images and adversarial examples, thereby avoiding any drop in clean image accuracy.
\end{abstract}


\section{Introduction} \label{intro}

\IEEEPARstart{T}{he} rapid development of deep neural networks (DNNs) has achieved great success in a variety of domains. However, the pervasive brittleness of DNNs, i.e., the vulnerability against the adversarial examples (AEs), which are the data with imperceptible perturbations added, has drawn significant attention~\cite{szegedy2013intriguing} because it raises doubts about DNNs being used in safety-critical applications in real-world scenarios.

While tremendous effort has been made to explore the mechanism of the adversarial examples, it is still an open problem. \cite{szegedy2013intriguing} suggests that adversarial examples are low-probability but dense statistical variation from the dataset. \cite{goodfellow2014explaining} argues that the linearity of deep models reduces their ability to resist adversarial examples. There are also works that try to explain the effect of adversarial examples by the property of the data itself instead of the training scheme. \cite{schmidt2018adversarially} shows that robustness requires a larger sample complexity, that is, a larger training set. More recently, Ilyas et. al.~\cite{ilyas2019adversarial} explain the phenomenon from the viewpoint of the inherent property of the data, that is, the fragility against adversarial examples comes from the non-robust (NR) features of the data. Because the model doesn't have any restrictions to use human-meaningful features to train the model, it may prone to rely on human-incomprehensible features to conduct the training. For natural images, these NR features are highly predictive while for adversarial examples, they could be easily perturbed and provides incorrect output. Correspondingly, the authors also demonstrate the existence of robust features that would preserve the robustness. 

\begin{figure}
\centering
  \includegraphics[width=\linewidth]{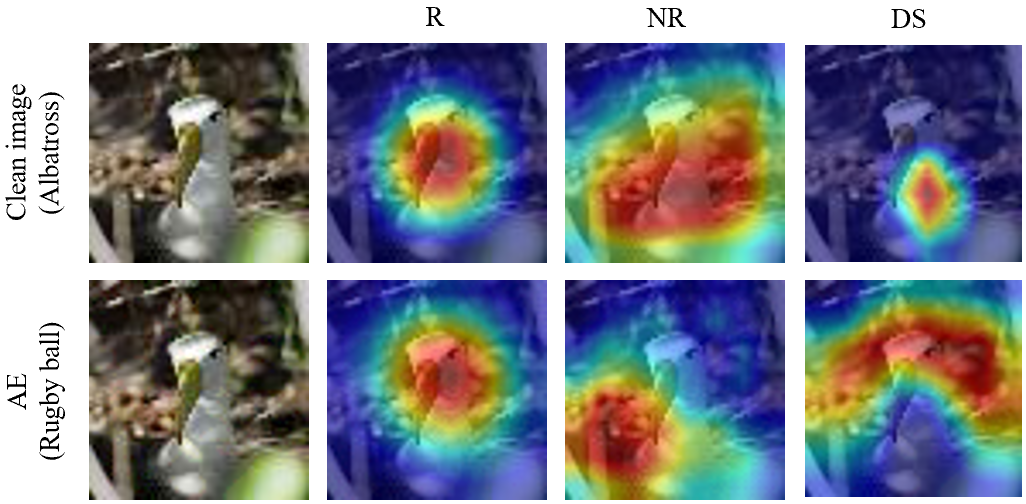}
  \caption{\revise{Grad-CAM~\cite{selvaraju2017grad} visualization of robust (R), non-robust (NR), and domain specific (DS) features of a natural image and its adversarial example.}}
  \label{fig:GradCAM}
\end{figure}

Motivated by~\cite{ilyas2019adversarial}, we strive to explore the robust features to improve the model's robustness. As a feature representation of an image is usually an entanglement of different types of features, we propose a feature disentanglement model to disentangle the robust features from other types of features. We argue there are three types of features in the raw features. The first one is the \textit{robust (R)} features which are the intrinsic representations of a specific class, such as shapes and colors which should be invariant to different domains, i.e., the natural (clean) images and adversarial examples if we treat them as different domains. As shown in the second column in Figure~\ref{fig:GradCAM}, the robust features remain unchanged for the clean image and its adversarial example. The second one is the \textit{non-robust (NR)} features which are some human-incomprehensible representations that are related to certain details. NR features from natural images and adversarial examples usually present different classification predictions as they can be easily contaminated by small perturbations. As shown in the third column in Figure~\ref{fig:GradCAM}, the non-robust features of the clean image, although different from robust features, focus on the regions around the object and may pay attention to some complementary information to robust features. In contrast, The non-robust features of the adversarial example focus on the regions far from the object and thus lead to an incorrect prediction. Both R and NR features are class-relevant but domain irrelevant. In addition to R/NR features, we also argue that there are other features relevant to the domains given the domain shift (e.g., the misalignment of statistic distribution and classification performance) between clean images and adversarial examples, we referred to them as \textit{domain specific (DS)} features. A discriminator is jointly trained in our framework to classify DS features into different domains.
We visualize the DS features in the last column in Figure \ref{fig:GradCAM}. It shows the focusing region of the domain discriminator for clean image and its adversarial example, which should have no association with the class label, indicating a lack of semantic meaning.
Once well-trained, the disentangled robust features from our framework can preserve a higher accuracy against adversarial attacks. Moreover, the adversarially trained models usually suffer from a performance drop for the natural images. \revise{For example, a naturally trained model can achieve approximately 96$\%$ testing accuracy on clean images while this number will drop to around 86$\%$ for a traditionally adversarial training model~\cite{madry2017towards}.} As a side product, the discriminator obtained from our framework performs outstandingly to tell whether a given image's domain specific features belong to a clean one or an adversarial example, and thus it can be applied for the adversarial example detection. For an image that is classified to be a natural image by our discriminator, it will be conducted by the model trained on natural training, thus, we can raise the testing accuracy on natural images to be almost the same as the natural-trained model.

There are recent studies trying to improve adversarial robustness through feature disentanglement or distillation. \cite{yang2021adversarial} splits features into class-specific and class-irrelevant ones, \cite{bashivan2021adversarial} utilizes an adversarial training strategy to extract the domain invariant features. 
\cite{yang2021class} decomposes an image into class-essential and class-redundant via a variational auto-encoder. \cite{zhou2021towards} also tries to find the attack-invariant features but it uses a pre-processing framework. None of them explicitly disentangle robust and non-robust features. \cite{kim2021distilling} extract robust by feature channel selection strategy and thus the extracted robust features are not purified. Our framework explicitly disentangles the R/NR and DS features and improves the adversarial robustness.

From the domain adaptation/generalization point of view, the natural images and the adversarial examples grow into two different domains because of the misalignment of statistic distribution and model performance~\cite{bashivan2021adversarial}. A natural question is that, while we can generate adversarial examples by one or more known attacks as source domains in training, is our model general enough to defend other unseen (in training) attacks (target domain)? This is a typical domain generalization problem~\cite{wang2021generalizing}. In the field of domain generalization, representation learning is a category of methods in order to learn domain-agnostic features that can generalize from source domain to unseen domain by either decomposition \cite{chattopadhyay2020learning} or generative modeling \cite{Peng2019DomainAL}, among which feature disentanglement provides an effective way to disentangle representation into domain-specific features and domain-invariant features \cite{wang2021generalizing, zhou2021domain}. Variants of feature-disentanglement frameworks may tackle the problem differently. \cite{Peng2019DomainAL} disentangle the representation into domain-invariant, domain-specific and class-irrelevant features. \cite{ilse2020diva} learns three latent spaces for domain, class and other variations independently. In this paper, we explicitly disentangle the domain-agnostic features (both R and NR), and further segregate the robust features that remain the prediction correct under different attacks. 

The extensive experiments on three widely used datasets, i.e., CIFAR-10, CIFAR-100 and Tiny-ImageNet, with different attacks demonstrate that robust features obtained from our model improve the model's adversarial robustness compared to the state-of-the-art approaches.

In summary, the contributions of this paper are as follows:
\begin{itemize}
\item We proposed a feature disentanglement based framework that can explicitly segregate robust features, that are invariant to the attacks, out of non-robust and domain-specific features.

\item The robust features show higher adversarial robustness than the state-of-the-art adversarial defense models.

\item As side products, the discriminator and domain-specific features can successfully distinguish between natural images and adversarial examples. With that, we can specify different classifiers for clean images and adversarial examples, and thus our model has almost no sacrifice on the accuracy of clean images.
\end{itemize}




\section{Related Works}\label{related}

\subsection{Adversarial Attacks.} Following the discovery of deep neural networks' vulnerability to adversarial examples, numerous research efforts have been dedicated to producing stronger adversarial attacks. \cite{szegedy2013intriguing} considers the generation of adversarial examples as a box-constrained optimization problem. \cite{goodfellow2014explaining} proposes a one-step attack named fast gradient sign method (FGSM) and leads out a lot of variants including basic iterative method (BIM) \cite{kurakin2016adversarial}, which is an iterative version of FGSM and MI-FGSM \cite{dong2018boosting} where momentum is integrated. In addition, the projected gradient descent (PGD) proposed by \cite{madry2017towards} and the CW attack proposed by \cite{carlini2017towards} are widely used as a criterion to measure the robustness of a deep model. \cite{moosavi2016deepfool} attempts to quantify the adversarial robustness by computing the minimal adversarial perturbation. In \cite{AA_atack_2020}, a per-sample attack named AutoAttack is introduced. AutoAttack is an ensemble of four attacks, APGD$_{CE}$, APGD$^T_{DLR}$, \cite{croce2020minimally} and \cite{andriushchenko2020square}. It attacks by computing the worst case for each image. In addition to additive attacks, there are also geometric attacks such as \cite{fawzi2015manitest, goodfellow2009measuring, kanbak2018geometric}, where geometric transformations are applied to the images to fool the classifier. Moreover, \cite{eykholt2018robust, huang2020universal, kurakin2016adversarial} tried to use patches in the physical world to confuse the model. Other than image classification, adversarial examples are crafted for other tasks such as objective detection \cite{wang2022daedalus}, deep retrieval models \cite{yang2020adversarial}, medical deep learning systems \cite{wang2022afeature} and power systems \cite{tian2022joint} to investigate the robustness of the models. In addition, adversarial attacks are also deployed to achieve further performance improvement \cite{liu2021point, shi2022adversarial}.

\subsection{Adversarial Defense.} Because of the discovery of adversarial examples, defense against the attacks emerges in response. Previous works prove that adversarial-training based defending methods are among the most effective and popular ones. Adversarial training utilizes adversarial examples to train the model, that is, the algorithm will first maximize the loss function to generate adversarial examples, and then minimize the loss function to update the parameters afterward. PGD attack is used in \cite{madry2017towards}, which can be considered as a standard adversarial training method. Based on this, \cite{kannan2018adversarial} adds constraints to pull the logits of natural images and their adversarial examples closer. \cite{Mao2019} and \cite{zhong2019adversarial} constrain the distance between feature vectors of natural images and their adversarial example with metric learning. \cite{shafahi2019adversarial} reduces the computational complexity of adversarial training by blending the generation of adversarial examples and the optimization of the model parameters. \revise{\cite{goldblum2020adversarially, zhu2021reliable} utilize knowledge distillation to transfer adversarial robustness to student networks} and \cite{Wang_2021_ICCV} applies knowledge distillation and bidirectional metric learning to correct the perturbations from adversarial examples for both feature maps and latent vectors. \cite{TRADES_2019} uses KL-divergence as a regularization to both generate the adversarial examples and train the model while \cite{zhang2019defense} uses feature scattering in latent space to accomplish the generation of attacks. \cite{yang2021class, wang2021revisiting} use information bottleneck to reduce redundant information and \cite{bashivan2021adversarial, zhou2021towards} try to find invariant features for better robustness. \cite{Wang2019} designs a bilateral adversarial training strategy. \cite{xie2019feature} denoises features to defend against adversarial attacks. In \cite{MART_2020}, the authors take the misclassified examples into consideration and further improve the robustness. A recent study \cite{shen2022adversarial} protects particular classes from adversarial attacks with the utilization of cost-sensitive classification and consequently achieves better average accuracy over all classes. Corresponding to attacks on object detection tasks, defending methods against adversarial attacks on object detection models are proposed \cite{li2020rosa}.

\subsection{Feature disentanglement.} Feature disentanglement is a popular branch in the field of domain generalization. \cite{Peng2019DomainAL} designed a domain-agnostic learning (DAL) framework to extract domain-invariant representations by a deep adversarial disentangled autoencoder (DADA) with domain-specific and class-irrelevant features as auxiliary. \cite{ilse2020diva, qiao2020learning} uses Variational Auto-Encoder and Wasserstein Auto-Encoder \cite{tolstikhin2017wasserstein} for disentanglement and generalization. \cite{chattopadhyay2020learning} devise domain specific masks to tackle domain generalization problems similarly as disentanglement does. In addition, similar architecture is used in domain adaptation \cite{lee2021dranet, chen2021dbin}, face presentation attack detection \cite{wang2020cross} and image-to-image translation \cite{lee2020drit++, song2022toward}.

\section{Proposed Method}

In this section, we introduce the process of accomplishing the disentanglement of robust, non-robust, and domain-specific features.

\begin{figure*}
  \centering
   \includegraphics[width=\linewidth]{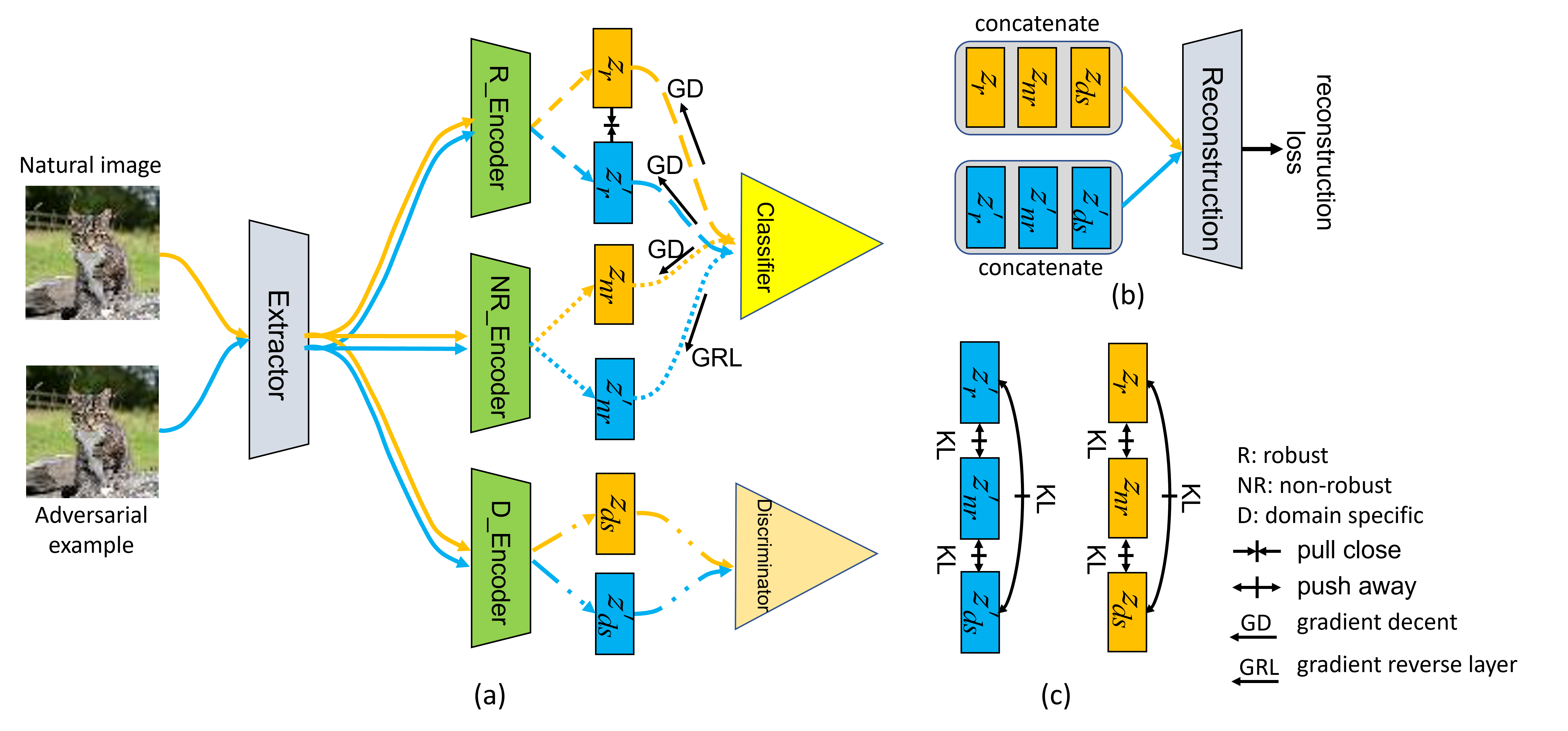}
  \caption{(a) Illustration of our proposed framework which segregates the robust features out of non-robust features and domain specific features by using a feature disentanglement model. (b) Reconstruction decoder to constraint disentanglement to not have information loss. (c) A further constraint between each pair of the three disentangled features makes them have small dependence on each other.}
  \label{fig:framework} 
\end{figure*}

\subsection{Preliminaries}

We first briefly introduce the standard adversarial training, including the adversarial examples generation and loss function. Let $F$ be a deep model parameterized by $\Theta$ and $\mathcal{D}$ be a dataset containing $N$ samples \revise{$\mathcal{D}=\left\{(\revise{x_i, y_i})\right\}_{i=1}^N$, where $y_i$ is the class label. We use $(x,y)$ to refer to an image-label pair for simplicity.} A standard way to generate adversarial examples is maximizing the classification loss as in Eq.~\ref{eq:ae}.
\begin{equation}\label{eq:ae} 
x_{adv} = \max \limits_{\delta \in\Delta} \mathcal{L}(F(x+\delta), y)
\end{equation}
$\delta$ refers to perturbation added to the image $x$ and $\Delta$ defines the constraint of an $\ell_{p}$-norm bound of the total perturbation, such as $\ell_{2}$ and $\ell_{\infty}$ norm. \revise{$\mathcal{L}$ is the loss function.}
Correspondingly, the adversarial training method optimizes the parameters of the model by those adversarial examples, thus can be written in a min-max manner as shown in Eq.~\ref{eq:mimmax}:
\begin{equation}\label{eq:mimmax} 
\min \limits_{\Theta} \mathop{\mathbb{E}} \limits_{x\in\mathcal{D}} \left[\max \limits_{\delta \in\Delta} \mathcal{L}(F(x+\delta), y)\right]
\end{equation}
where \revise{$\mathbb{E}[\cdot]$ refers to the expectation over the whole dataset.} The outer minimization in Eq.~\ref{eq:mimmax} is to train the target model on the adversarial examples, while the inner maximization is to generate strong adversarial examples. Currently, the most popular attack is the Projected Gradient Descent (PGD)~\cite{madry2017towards}, which utilizes an iterative way to obtain the adversarial examples:
\begin{equation}\label{eq:ae_it} 
x_{adv}^{t+1} = x_{adv}^t + \alpha \cdot \nabla_{x} \mathcal{L}(F(x_{adv}^t), y), \,\,\,\,\,st. \,\,\,\|x_{adv}-x\|_{\infty}\leq\epsilon
\end{equation}
where $t\in\left\{1,2,\dots,T\right\}$ is the number of steps, $\alpha$ is the step size. The constraint ensures the perturbation strength of the adversarial example is within a $\epsilon$-ball, which is also called attack budget, around $x$ to ensure perceptual similarity.

\subsection{Feature Disentanglement}

There is a distribution shift between the natural (original) images and adversarial examples because of the add-on adversarial perturbations. Therefore, we consider the natural images and adversarial examples as two domains.
We use $x_{adv} \in \mathcal{D}'$ to denote images from the adversarial domain and $x \in \mathcal{D}$ for the images from the natural domain.

As shown in Figure~\ref{fig:framework}, an input image is first fed into the feature extractor $G$ parameterized by $\theta$ to get the intermediate features, referred to as $f$, $f'$ for $x$ and $x_{adv}$, respectively. 
\begin{equation}\label{eq:ext} 
f = G_{\theta}(x); f' = G_{\theta}(x_{adv})
\end{equation}
The intermediate features, both of $f$ and $f'$, are the entanglements of robust, non-robust and domain-specific information. We then use three encoders, i.e., robust encoder $E_{\omega_r}$, non-robust encoder $E_{\omega_{nr}}$ and domain-specific encoder $E_{\omega_{ds}}$ to disentangle them explicitly. $f$ (or $f'$) is fed into these three encoders and expected to output disentangled robust features $z_r$ ($z'_r$), non-robust features $z_{nr}$ ($z'_{nr}$) and domain specific features $z_{ds}$ ($z'_{ds}$), respectively, denoted as:
\begin{equation}\label{eq:enc} 
\widetilde{z}_{i} = E_{\omega_{i}}(\widetilde{f})
\end{equation}
where $i\in\{r, nr, ds\}$, $\widetilde{z}_{i} \in \{{z}_{i}, {z'}_{i}\}$ and $\widetilde{f} \in \{f, f'\}$. 

A natural constraint for the complete disentanglement is that the disentangled features $z_r, z_{nr}, z_{ds}$ ($z'_r, z'_{nr}, z'_{ds}$), should have the small dependence between each other as possible. We utilize a Kullback-Leibler (KL) divergence loss $\mathcal{L}_{kl}$ between each pair of the three disentangled features, for natural images and adversarial examples, respectively, as shown in Figure~\ref{fig:framework} (c).

\textbf{Robust (R) features.} R features are the intrinsic representations of a specific class and they should be robust to the perturbations, i.e., invariant between natural images and adversarial examples. Therefore, the robust features of the natural image and its adversarial example should be as similar as possible. We minimize the loss of angular distance between these two features as followed:
\begin{equation}\label{eq:cos_dist} 
\mathcal{L}_{dist}(\theta,\omega_{r})= 1-\frac{\left|z_r \cdot z'_r \right|}{\|z_r\|_2 \|z'_r\|_2}
\end{equation} 
R features are what we rely on for our task, they should have good classification ability. We impose the classification loss (specifically the cross-entropy) on them, as shown in Eq.~\ref{eq:ce}.

\textbf{Non-robust (NR) features.} NR features are some human-incomprehensible representations that are related to certain details of an object which can be easily contaminated by small perturbations and thus makes natural images and adversarial examples have different classification predictions. NR features in a natural image, $z_{nr}$, should still make a correct prediction, while in an adversarial example, $z'_{nr}$, they should make an incorrect prediction. We also impose the cross-entropy loss on them. The following equation absorbs both R and NR features:
\begin{equation}\label{eq:ce} 
\small
\mathcal{L}_{ce}(\theta,\omega_{\{r, nr\}},\phi)=
-\mathop{\mathbb{E}} \limits_{z\in\{z_r, z_{nr},z'_r,z'_{nr}\}} \sum_{k=1}^C\mathds{1}{[k=y_t]}\log(C_{\phi}(z))
\end{equation}
where $C_{\phi}(z)$ is out target classifier. As shown in Figure~\ref{fig:framework} (a), all of R and NR features branches utilize gradient descent (GD) for updating the model's weights except the $z'_{nr}$ branch as it makes the model output an incorrect prediction. Therefore, we update the weights of non-robust encoder $\omega_{nr}$ in the direction of increasing the cross-entropy loss, similar to that in gradient reversal layer (GRL) \cite{ganin2016domain}. Specifically, $-\lambda$ is multiplied to the gradients of $\omega_{nr}$ only for optimization during backpropagation i.e. using $-\lambda \frac{\partial \mathcal{L}_{ce}}{\partial \omega_{nr}}$ instead of $\frac{\partial \mathcal{L}_{ce}}{\partial \omega_{nr}}$.

\textbf{Domain-specific (DS) features.} R/NR features are all class relevant features. Due to the misalignment of statistic distribution and classification performance) between clean images and adversarial examples, there must exist some domain-specific features that represent such domain shift. The domain-specific features should capture the characteristics of the natural images and adversarial examples and thus the domain-specific features from different domains could be identified by a jointly trained discriminator $D_\psi$. The domain-specific encoder $E_{\omega_{ds}}$ and $D_\psi$ are trained adversarially~\cite{yang2021adversarial}. The discriminator is first trained to distinguish samples from different domains:
\begin{equation}\label{eq:bce} 
\begin{split}
\mathcal{L}_{bce}(\psi)=&-\mathop{\mathbb{E}} \limits_{z\in\{z_r,z_{nr}, z_{ds}\}} \log(D_\psi (z))\\
&-\mathop{\mathbb{E}} \limits_{z\in\{z'_r,z'_{nr}, z'_{ds}\}} \log(1-D_\psi (z))
\end{split}
\end{equation}
We then fix the discriminator and update three encoders by optimizing the adversarial loss in Eq.~\ref{eq:discriminator}. Specifically, the R and NR encoders are trained to fool the discriminator by minimizing the adversarial loss as they are domain invariant features, while the DS encoder is trained to be classified by the discriminator.
\begin{equation}\label{eq:discriminator} 
\small
\begin{split}
&\mathcal{L}_{adv}(\omega_{\{r, nr\}})=\mathop{\mathbb{E}} \limits_{z\in\{z_r,z_{nr}\}} [\log(D_\psi( z))]
+\mathop{\mathbb{E}} \limits_{z\in\{z'_r,z'_{nr}\}}[\log(1-D_\psi (z)]\\
&\mathcal{L}_{adv}(\omega_{ds})=-\mathop{\mathbb{E}}[\log(D_\psi(z_{ds}))]-\mathop{\mathbb{E}} [\log(1-D_\psi (z_{ds})]
\end{split}
\end{equation}

\textbf{Reconstruction.} While we disentangle the features, we don't want to have any information loss. Therefore, we constraint the disentangled features $\widetilde{z}_r, \widetilde{z}_{nr}, \widetilde{z}_{ds}$ to be able to recover the original features $\widetilde{f}$, as shown in Figure~\ref{fig:framework} (b).
\begin{equation}\label{eq:reconstruction} 
\mathcal{L}_{res}(\theta_{rec}, \omega_{\{r, nr, ds\}})=\mathop{\mathbb{E}} \limits_{\widetilde{z}\in\{z,z'\}} \|R_{\theta_{rec}}([\widetilde{z}_r, \widetilde{z}_{nr}, \widetilde{z}_{ds}])-\widetilde{f}\|_1
\end{equation}
where $\theta_{rec}$ is the parameter of a reconstruction module.

\textbf{Domain diversify.} The diversity of training data is of great importance to a model’s generalization ability~\cite{wang2021generalizing}. We try to obtain a defense model trained on PGD attacks that will be general enough to defend against other unseen (in training) attacks. Therefore, we diversify the training domain by randomly selecting the number of steps, as well as the step size of the PGD attack. With that, we expect our model to have stronger generalizability.

The overall procedure of our proposed model is shown in Algorithm~\ref{alg:alg}. 

\begin{algorithm}
  \caption{Robust Feature Disentanglement.}
  \label{alg:alg}
  \KwIn{Clean image set $\mathcal{D}$, epoch number $N$, batch size $b$, learning rate $\gamma$}

  \KwOut{Network parameter $\theta$, $\omega_{\{r,nr,ds\}}$, $\phi$, $\psi$, $\theta_{rec}$}
  
  \For{{\em epoch} $= 1, ..., N$}
  {
    \For{{\em minibatch} $\{x^i, y^i\}^b_{i=1}$ } 
    {
      a. for each image $x^i$, generate its adversarial examples $x^i_{adv}$ using Eq. \ref{eq:ae_it} \;
      b. feed $x^i$ and $x^i_{adv}$ into the feature extractor and three encoders to get disentangled features using Eq. \ref{eq:ext} and \ref{eq:enc}\;
      c. optimize $\omega_{\{r,nr,ds\}}$ by pair-wise $\mathcal{L}_{kl}$ losses of disentangled features, as shown in Fig.~\ref{fig:framework} (c)\;
      d. optimize $\omega_{r, nr}$ with Eq. \ref{eq:ce} for classification ability \;
      e. optimize $\theta$, $\omega_r$ with Eq. \ref{eq:cos_dist} to align the robust features from $x^i$ and $x^i_{adv}$ \;
      f. optimize $\theta$, $\omega_{\{r,nr,ds\}}$, $\phi$, $\psi$ with Eq. \ref{eq:bce} and Eq. \ref{eq:discriminator} to train the discriminator and accomplish domain disentanglement \;
      g. optimize $\omega_{\{r,nr,ds\}}$, $\theta_{rec}$ for reconstruction.
    }
  }
   return  $\theta$, $\omega_{\{r,nr,ds\}}$, $\phi$, $\psi$, $\theta_{rec}$\;
\end{algorithm}

\section{Experimental Results}

\subsection{Experimental settings} \label{sec:experimentsettings}
\textbf{Dataset} The performance of our model is evaluated on three popular datasets: CIFAR-10, CIFAR-100 and Tiny-ImageNet.
\textit{CIFAR-10} contains 60k images (the image size is $32\times32$) grouped into 10 classes. 50k images for training and 10k images for testing. \textit{SVHN} dataset, for the Street View House Number. The training set of SVHN contains 73257 images and the testing set has 26032 images belonging to 10 classes. \textit{CIFAR-100} is similar to CIFAR-10 except that it consists of 100 classes and 600 images for each class. \textit{Tiny-ImageNet} is a subset of ImageNet, which contains 200 classes. For each class, there are 500 training images and 50 validation images. The resolution of images in Tiny ImageNet is $64\times64$.

\textbf{Comparison methods} We compare the performance of our model with the following methods: (1) {\em undefended model (UM)}, which is trained by the clean image set only; (2) {\em adversarial training (AT)}~\cite{madry2017towards}, where the model is trained by adversarial examples generated from PGD attack; (3) {\em TRADES \cite{TRADES_2019}} and (4) {\em MART \cite{MART_2020}}, two defending methods with different regularization terms. (5) {\em LBGAT \cite{cui2021learnable}}, which improve robustness by boundary guidance of undefended model (65) {\em DRRDN\cite{yang2021adversarial}} and (7) {\em AFD-WGAN\cite{bashivan2021adversarial}}, which uses feature disentanglement and desensitization to defend adversarial attacks, \revise{(8) IAD \cite{zhu2021reliable}, which utilizes knowledge distillation to transfer robustness. For IAD, we adopt the results with AT as the teacher model here.} We use FGSM~\cite{goodfellow2014explaining}, PGD~\cite{madry2017towards}, CW~\cite{carlini2017towards} attacks to evaluate the models. We also test the models on black-box attack and AutoAttack (AA)~\cite{AA_atack_2020}, which is a parameter-free attack that ensembles four diverse attacks.

\textbf{Implementation details} We use the residual network as the backbone of our model. The feature extractor $G_\theta$ consists of two residual blocks, \revise{each contains five basic residual blocks where two $3\times3$ convolutional layers are included}. The encoder is a CNN architecture with one  residual block, while the reconstruction module is composed of transposed convolutional layers. The fully connected layer is used as a classifier. We utilize a two-layer fully connected network as the discriminator. For all the datasets, the initial learning rate is 0.1 for the adversarial training of $G_\theta$, $E_{\omega_r}$ and $C_\phi$ and other components have a smaller learning rate. We utilize SGD optimizer and schedule the learning rate decay at 100, 105 and 110 epochs. The batch size is 128 and we use early stopping as suggested by \cite{rice2020overfitting}. In order to increase the domain diversity, the attack budget $\epsilon$ is randomly selected between 8 and 12 for each batch. The number of steps is randomly sampled from $\left[8, 16, 24, 32\right]$ and step size ranges from 2 to 4. 

\subsection{Adversarial robustness against different attacks}
We evaluate the accuracy of our model against various attacks. \revise{We conducted five runs of our model across all settings and reported the mean and standard deviations resulting from these runs.} In Table~\ref{tab:cifar10_result}, we show the results on CIFAR-10. The accuracy of our model on clean images in the table is directly from the robust features through the classifier. With the help of the well-trained discriminator, this accuracy can be improved to almost the same accuracy as the undefended model, i.e., around $96\%$, we will describe this in section~\ref{sec:clean}. The rest of Table~\ref{tab:cifar10_result} includes accuracy against both white-box attacks and black-box attacks. \textbf{White-box attack} denotes the adversary has full access to the target model, including the architecture and parameters. Our model outperforms comparison methods under most of the white-box attacks. Especially, our model shows the best performance on AA attacks which is known as the most challenging defense task. \textbf{Black-box attack} denotes there is no information known about the target model. We trained a surrogate model with only clean images and generate adversarial examples from this model to attack our model. As shown in Table \ref{tab:cifar10_result}, our model still keeps a high accuracy against black-box attacks. \revise{To gain a more comprehensive understanding of the model's performance under different adversarial scenarios during training, we have expanded our evaluation by incorporating results from training the model with attacks other than PGD. We specifically consider the CW attack and an adaptive version of the PGD attack, i.e., the difference of logits ratio (DLR)~\cite{AA_atack_2020}. The results show that our proposed model performs quite well when trained under different attacks.}

\begin{table*}[htbp]\small
\centering\small
\caption{Evaluation results on CIFAR-10 under different attacks. ``BB" denotes the black-box attack. The best accuracy is highlighted. We also present the results of our model when training with different attacks, namely CW and the difference of logits ratio (DLR).} \label{tab:cifar10_result} 
\begin{tabular}{c|cccccc}
\hline \hline
Methods & clean & FGSM & PGD$_\infty$  & CW$_\infty$  & AA & PGD(BB) \\
\hline 
UM & \textbf{96.20} & 31.39 & 0 & 0 & 0 & - \\
AT~\cite{madry2017towards} & 86.19  &  62.42  &  45.57  & 46.26 &  44.04 & 79.55 \\
TRADES~\cite{TRADES_2019} & 83.50   & 63.68  & 52.80    & 50.90 &   52.70 & 81.60 \\ 
MART~\cite{MART_2020} & 82.16 & 63.91  & 52.67  & 49.44 &   51.20  & 82.60 \\ 
LBGAT~\cite{cui2021learnable} & 88.22  & -   &  54.66    & 54.29 &   52.23  & - \\ 
DRRDN~\cite{yang2021adversarial} & 85.76 & 62.81 & 52.32 & 51.88 & - & 82.65 \\
AFD-WGAN~\cite{bashivan2021adversarial} & 85.95 & - & \textbf{59.38}   & - & 37.33 &  \textbf{84.74} \\
\revise{IAD~\cite{zhu2021reliable}} & \revise{85.09} & \revise{66.54} & \revise{55.45} & \revise{54.63} & \revise{52.29} & \revise{-}\\
Ours & 83.52$\pm$0.12    & \textbf{66.81}$\pm$\textbf{0.79}  & 57.97$\pm$0.26   & \textbf{55.39}$\pm$\textbf{0.25}  &  \textbf{53.64}$\pm$\textbf{0.22} & 83.33$\pm$0.21 \\
\hline
\revise{Ours-CW} & 80.17$\pm$0.11   & 65.51$\pm$0.35  & 57.79$\pm$0.30   & 56.02$\pm$0.18  &  51.60$\pm$0.11 & 80.03$\pm$0.15  \\
\revise{Ours-DLR~\cite{AA_atack_2020}} & 87.20$\pm$0.12    & 69.61$\pm$0.27  & 63.28$\pm$0.27   & 50.98$\pm$0.29  &  48.30$\pm$0.36 & 87.03$\pm$0.08 \\
\hline \hline
\end{tabular}
\end{table*}

The results on the SVHN dataset are shown in Table~\ref{tab:svhn_result}. \revise{For SVHN, the UM model has a success rate lower than $1\%$ against PGD and CW attacks. The ML method is the best among the comparison models which can achieve over $51\%$ for both attacks.} Our model outperforms the comparison methods on all attacks by a large margin \revise{and the accuracy could reach $55.20\%$ and $54.09\%$ for PGD and CW attacks}.

\begin{table}[htbp]\small
\centering
\caption{Evaluation results on SVHN, under different widely used attacks.} \label{tab:svhn_result}
\begin{adjustbox}{width=.5\textwidth}
\begin{tabular}{c|cccc}
\hline \hline
 Attacks & clean & FGSM &  PGD$_{\infty}$  & CW$_{\infty}$\\
\hline 
UM & \textbf{96.36$\%$}  & 46.33  & 0.33  & 0.37  \\
AT
& 91.55 & 67.13  & 45.64  & 47.14 \\
ML
& 83.95 & 70.28 & 51.91 & 51.25 \\
TRADES
& 91.16 & 69.85 & 50.90 & 50.85 \\
MART
& 91.16 & 67.31 & 48.72  & 50.52 \\
\hline
Ours & 91.01$\pm$0.73 & \textbf{72.41}$\pm$\textbf{1.10} & \textbf{55.20}$\pm$\textbf{0.52} & \textbf{54.09}$\pm$\textbf{0.66}  \\
\hline \hline
\end{tabular}
\end{adjustbox}
\end{table}

The results on CIFAR-100 and Tiny ImageNet are shown in Tables \ref{tab:result_cifar100}. \revise{For CIFAR-100, the undefended model (UM) misclassifies almost all the adversarial examples generated with strong attacks such as PGD and AA with only $0.01\%$ success rate for PGD and $0.02\%$ for AA. AT can increase the accuracy against PGD attack to $18.54\%$ while methods such as TRADES and AFD-WGAN can further improve the number. Our model outperforms other methods on PGD attack with accuracy $31.78\%$. For Tiny ImageNet, it's hard for UM model to make correct classifications against adversarial attacks. The comparison methods can improve the accuracy to over $10\%$ against PGD and AA attacks to the best. Our method greatly promoted these numbers to $24.78\%$ and $20.64\%$ separately.} Moreover, although our model is trained on $\ell_\infty$-norm attacks, it surpasses other comparison methods on $\ell_2$-norm attacks such as CW$_2$ by a large margin, showing superior generalization ability of our model.

\begin{table}[htbp]\small
\centering
\caption{ Evaluation results on CIFAR-100 and Tiny-ImageNet datasets under different attacks. The best accuracy in each dataset is highlighted.} \label{tab:result_cifar100}
\begin{adjustbox}{width=0.5\textwidth}
\begin{tabular}{c|c|cccc}
\hline \hline
& Methods & clean & PGD$_\infty$  & CW$_2$  & AA  \\
\hline 
\parbox[t]{2mm}{\multirow{6}{*}{\rotatebox[origin=c]{90}{CIFAR-100}}}  &  UM & \textbf{76.76} &  0.01  & 0.52  & 0.02   \\
&AT
& 56.49 &  18.54  & 17.71  & 18.30   \\ 
&TRADES
&  60.32 &  25.11  &  20.52 &  21.10  \\ 
&LBGAT
& 60.64 & 30.56 & - & \textbf{29.33}  \\
&AFD-WGAN
& 58.87  &  22.35  & \textbf{25.33} &  18.00   \\
\cline{2-6}
&Ours & 56.64$\pm$0.57  & \textbf{31.78}$\pm$\textbf{0.46}  & 24.31$\pm$0.10 &  27.93$\pm$0.21   \\
\hline \hline
\parbox[t]{2mm}{\multirow{5}{*}{\rotatebox[origin=c]{90}{TI}}} & UM &  \textbf{64.62} & 0.07 &  0   &  0 \\
&AT
& 43.80 & 12.62 &  4.90   &  9.48 \\ 
&TRADES
& 37.70 & 13.26 &  4.11   &  12.57  \\ 
&AFD-WGAN
& 47.70 & 11.49 &  5.90  & 9.45 \\
\cline{2-6}
&Ours   & 46.54$\pm$0.49 &  \textbf{24.78}$\pm$\textbf{0.18} &  \textbf{28.29}$\pm$\textbf{0.32}  & \textbf{20.64}$\pm$\textbf{0.16}  \\
\hline \hline
\end{tabular}
\end{adjustbox}
\end{table}

\subsection{\revise{Evaluation of additional network architectures on Imagenette.}}

\revise{In this section, we aim to evaluate the scalability of our method in handling larger-sized images and its ability to generalize across different neural network architectures. To achieve this, we perform experiments on the Imagenette dataset, which contains larger images, using three neural network architectures: Inception v4 (Inc-v4) \cite{szegedy2017inception}, Inception ResNet v2 (IncRes-v2) \cite{szegedy2017inception}, and ResNet v2-152 (Res-v2-152) \cite{he2016identity}. Imagenette is a subset of ImageNet, encompassing ten classes including tench, English springer, cassette player, chain saw, church, French horn, garbage truck, gas pump, golf ball, and parachute. The images of each class in the Imagenette dataset are identical to the original ImageNet dataset with an average image size of $469\times387\times3$ pixels.}

\revise{For the training on the Imagenette dataset, we used a learning rate of 0.01 following \cite{dong2018boosting}. The parameter numbers for Inc-v4, IncRes-v2, and Res-v2-152 are approximately 43M, 56M, and 60M, respectively. Therefore, for Inc-v4, the batch size remained at 128, whereas for IncRes-v2 and Res-v2-152, the batch sizes were reduced to 64 and 40, respectively, due to GPU limitations. All the comparison methods for each architecture use the same batch size for a fair comparison.}

\revise{The results for Imagenette dataset are presented in Table \ref{tab:imagenette}. Our model outperforms other methods under all the architectures. For Inc-v4 and IncRes-v2, our model surpasses other methods by a large margin. Specifically for Inc-v4, our method has imporved the accuracy by $7.33\%$ (from $30.68\%$ to $38.01\%$) and for IncRes-v2, The accuracy has $5.57\%$ increment (from $36.92\%$ to $42.49\%$).}

\begin{table*}[ht]
\caption{Evaluation and comparison of other neural network architectures on Imagenette dataset (without pretraining).}\label{tab:imagenette} 
\centering\small
\begin{tabular}{c|cc|cc|cc}
\hline \hline
\multirow{2}{*}{Models} & \multicolumn{2}{c|}{\revise{Inc-v4}}  & \multicolumn{2}{c|}{\revise{IncRes-v2 (bs=64)}} & \multicolumn{2}{c}{\revise{Res-v2-152 (bs=40)}}\\
\cline{2-7}
& clean & PGD$_{\infty}$  & clean & PGD$_{\infty}$  & clean & PGD$_{\infty}$ \\
\hline 
AT
& \textbf{80.71} & 21.76 & \textbf{87.69} & 24.38 & \textbf{87.95} & 27.34 \\
MART
& 28.31 & 21.94 & 52.64 & 35.44 & 64.05 & 42.75\\
TRADES
& 68.82 & 30.68 & 74.98 & 36.92 & 70.19 & 32.48 \\
Ours & 66.93$\pm$0.95 & \textbf{38.01}$\pm$\textbf{0.48} & 76.71$\pm$0.24 & \textbf{42.49}$\pm$\textbf{.27}& 73.88$\pm$0.21 & \textbf{43.36}$\pm$\textbf{0.19} \\
\hline \hline
\end{tabular}
\end{table*}

\subsection{Evaluation of k-Nearest Neighbor (k-NN) classifier}\label{sec:knn}

\begin{table*}[ht]
\caption{Evaluation and comparison between k-NN and softmax classifiers (k-NN/softmax).}\label{tab:knn}
\centering\small
\begin{tabular}{c|cc|cc|cc}
\hline \hline
\multirow{2}{*}{Models}&\multicolumn{2}{c|}{CIFAR-10} & \multicolumn{2}{c|}{SVHN} & \multicolumn{2}{c}{Tiny ImageNet}\\
\cline{2-7}
& clean & PGD$_{\infty}$  & clean & PGD$_{\infty}$  & clean & PGD$_{\infty}$  \\
\hline 
AT
& 87.1  / 86.2  & 47.5  / 45.6  & 91.5  / 91.6   & 45.8  / 45.6  & 36.6  / 42.3  & 20.2  / 19.6 \\
ALP
& 89.6  / 89.8  & 48.9  / 48.5   & 91.4  / 91.3  & 52.0  / 52.2  & 35.2  / 41.5  & 20.3  / 20.0 \\
ML
& 86.3  / 86.2  & 51.7  / 51.6  & 84.3  / 84.0  & 52.0  / 51.9  & 34.0  / 40.6 & 20.7  / 20.7 \\
Ours & 83.0 / 83.3 & 58.1 / 58.0 &  92.6 / 90.4   & 55.1 / 55.0 & 42.0 / 46.2 & 22.6 / 24.6\\
\hline \hline
\end{tabular}
\end{table*}

We use k-Nearest Neighbor (k-NN) method as a substitute of the original (softmax) classifier in the model following~\cite{Mao2019}. In our experiment, \revise{we use the same settings as \cite{Mao2019} where} $k$ is set to 50 and the features from the penultimate layer are used to build the k-NN classifier. 

The results for both k-NN classifier and the original classifier are shown in Table~\ref{tab:knn} on three datasets, CIFAR-10, SVHN and Tiny ImageNet. The comparison methods include AT~\cite{madry2017towards} ALP~\cite{kannan2018} SML~\cite{Mao2019}. Among the four approaches, our method can obtain a consistently higher accuracy against adversarial attacks on all datasets. Besides, although the k-NN is one of the simplest classifiers, it can achieve similar results compared with the original classifier. These quantitative results, coupled with the visualization illustrations in Section~\ref{sec:tsne}, demonstrate the features obtained by our model have better distribution in latent space. That is, the adversarial robustness of our model is from the latent space, not the classifier itself.

\subsection{Adversarial example detection and a simple strategy to improve accuracy on clean image}\label{sec:clean}
As a side product, the discriminator and domain specific features have great capability to distinguish the clean image and adversarial example, which can be used for adversarial example detection. Built upon that, we propose to use the natural trained model for clean images and our model for adversarial examples, to improve the accuracy of clean images. Our model can achieve $99.86\%$ and $100\%$ for true positive rate and true negative rate, respectively, in adversarial example detection, which makes a significant improvement on clean images (from $83.52\%$ to $96.20\%$), with a negligible drop on adversarial examples (form $57.97\%$ to $57.88\%$).

\subsection{Visualization of Domain Specific Features}\label{DS_vis}
\begin{figure}[h]
  \centering
   \includegraphics[width=.75\linewidth]{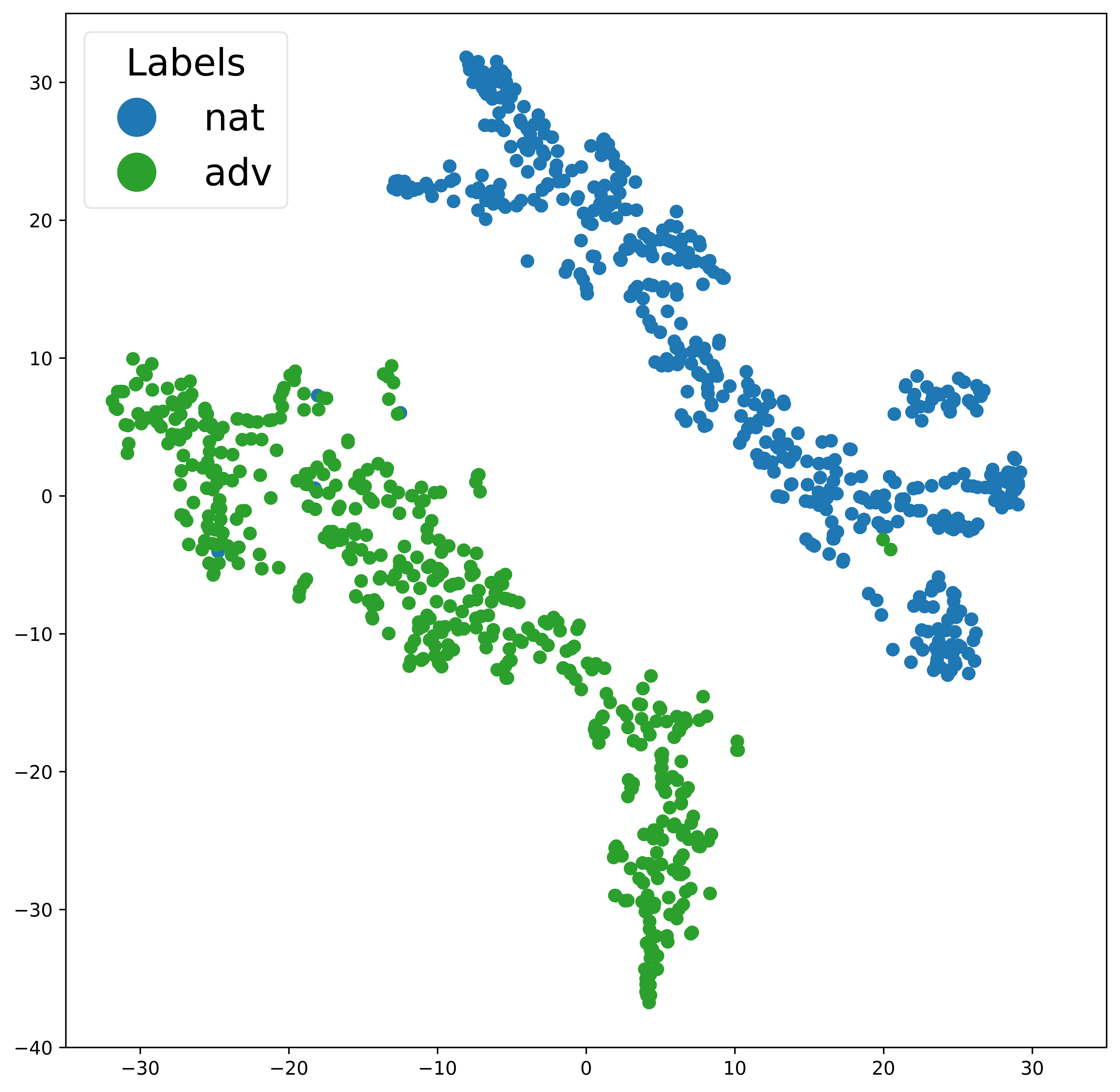}
  \caption{Domain specific features distributions for natural images (nat) and their adversarial examples (adv).}
  \label{fig:ds_tsne} 
\end{figure}

We visualize the distribution of the domain specific (DS) features for 500 randomly selected samples and their adversarial examples, using t-SNE, in Figure~\ref{fig:ds_tsne}. We can see that the DS features of natural images and adversarial examples are quite discriminable even in 2-dimensional space, while there are very few mis-clustering samples, we believe that they should be more easily classified in original high dimensional space.

\begin{figure}[ht]
  \centering
   \includegraphics[width=\linewidth]{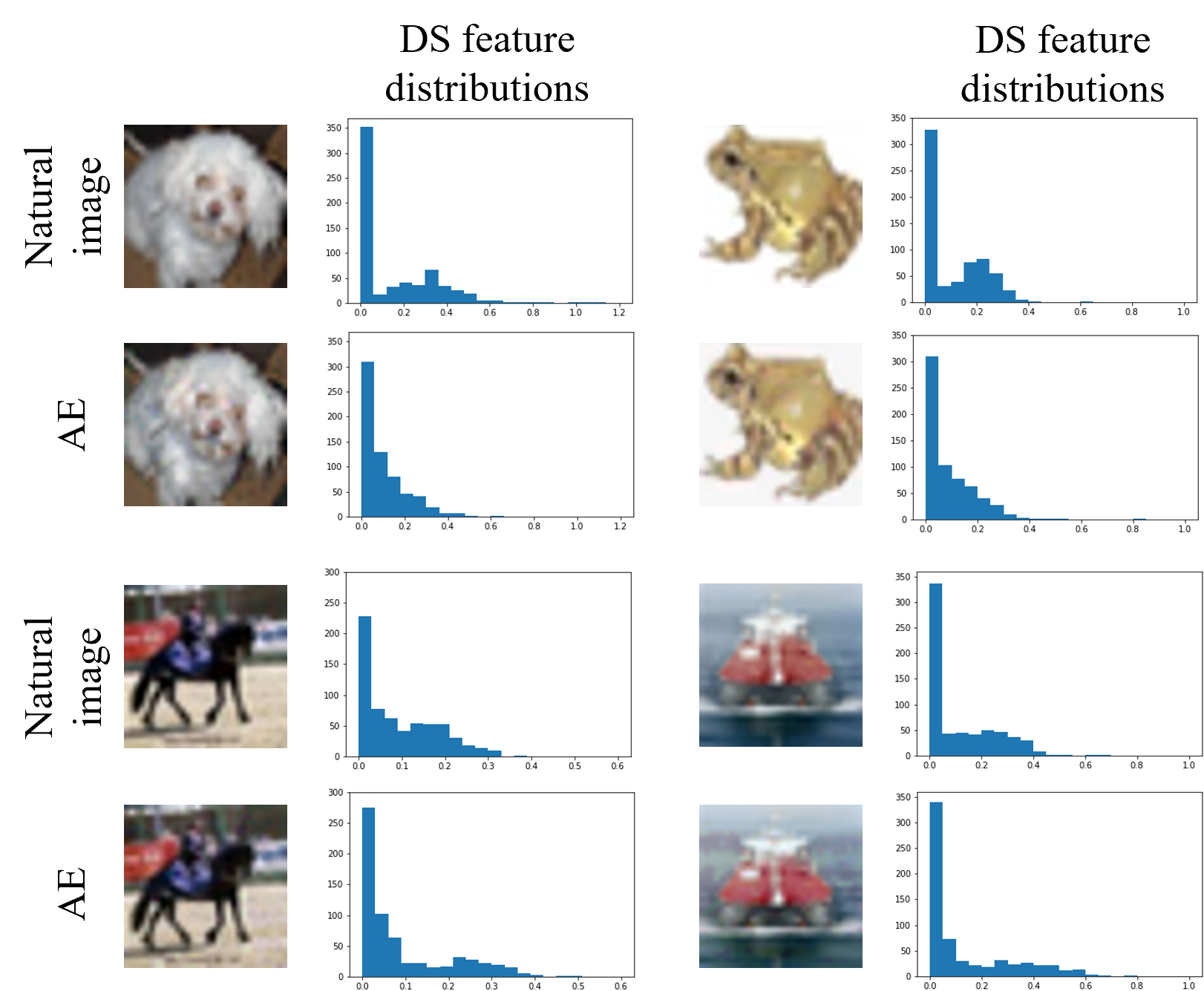}
  \caption{Illustrations of Domain specific features distribution for natural images and their adversarial examples (AE).}
  \label{fig:ds_feature} 
\end{figure}

We visualize the domain specific (DS) features, a 640-dimension vector that is obtained after the average pooling of a 640-channel feature maps output from the domain specific encoder, for natural images and their adversarial examples. Figure~\ref{fig:ds_feature} illustrates the DS feature histograms, which show the intensity distributions of the features, of some sampled natural images and their adversarial examples. 
From the figure, we can see that the intensity distributions for natural images and adversarial examples are quite different, which means the DS encoder has the ability to discriminate the features from natural images and adversarial examples. 

\subsection{\revise{Visualization of robust and non-robust Features}}\label{r_nr_vis}

\begin{figure}[h]
  \centering
   \includegraphics[width=.75\linewidth]{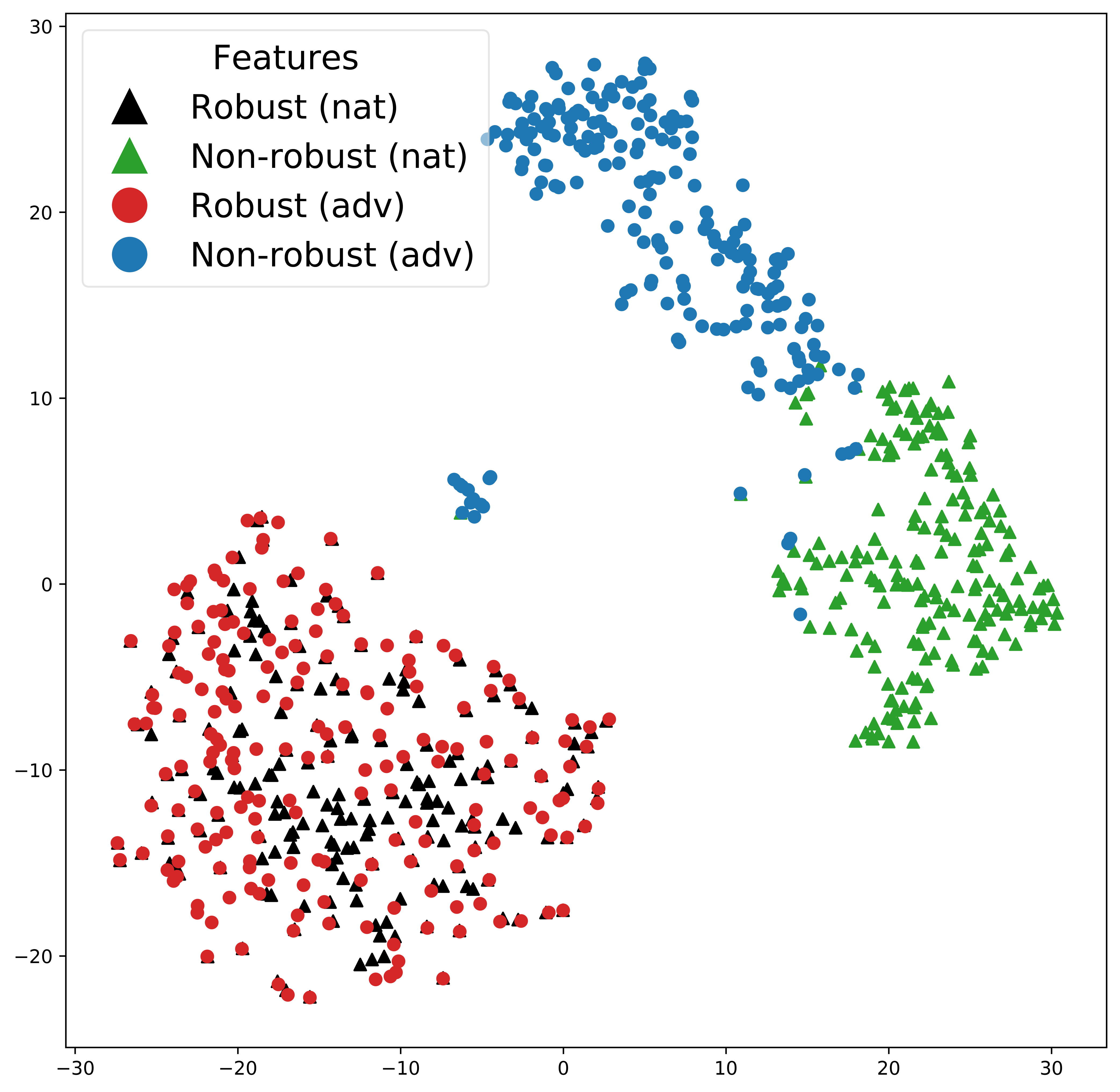}
  \caption{Robust and non-robust feature distributions for natural images (nat) and adversarial examples (adv).}
  \label{fig:r_nr_tsne} 
\end{figure}

\revise{In Figure \ref{fig:r_nr_tsne}, we visualize the distributions of the robust and non-robust features of natural images and their corresponding adversarial examples. The natural images are from the same class (frog) and the AEs are generated with PGD attack. As in Figure \ref{fig:r_nr_tsne},  our method successfully disentangles the robust and non-robust features. The robust features extracted from both natural images and adversarial examples are positioned within the same region. This alignment is due to the explicit pulling together achieved by the angular loss Eq. 6. }

\revise{On the other hand, the non-robust features extracted from natural images and adversarial examples are visibly separated. These features are the components modified by the adversarial perturbations to decrease (natural images) and increase (adversarial examples) the classification cross-entropy loss Eq. 7 and are effectively disentangled by our proposed method. The visualizations in Figure \ref{fig:r_nr_tsne} provide valuable insights into the disentanglement process, demonstrating the efficacy of our approach in capturing both robust and non-robust features for adversarial examples.}

\subsection{Different attack iterations}
\begin{figure}
\centering
  \includegraphics[width=.8\linewidth]{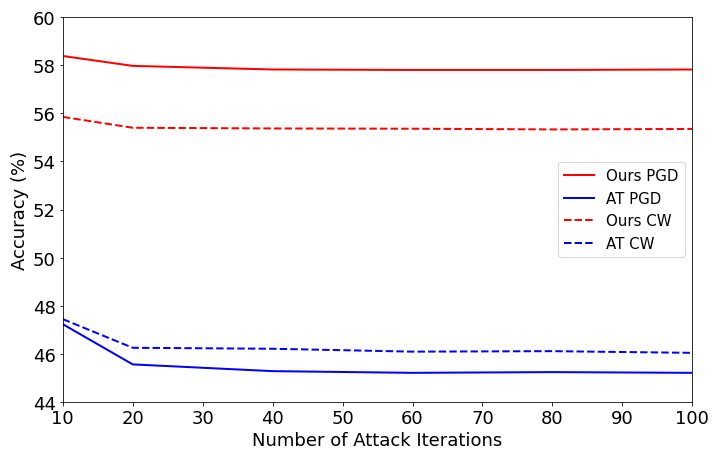}
  \caption{The accuracy under different attack iterations}
  \label{fig:it}
\end{figure}
We evaluate the model's robustness under different attack iterations, from 10 to 100. Figure~\ref{fig:it} shows that our model consistently outperforms standard AT, while it is not sensitive to the attack iterations although the accuracy monotonically declines along with the iteration number increase, for example, our model achieves $58.38\%$, $57.97\%$, and $57.82\%$ for 10, 20, and 100 iterations of PGD attack, respectively


\subsection{Visualization in feature space}\label{sec:tsne}
\begin{figure*}
  \centering
   \includegraphics[width=.8\linewidth]{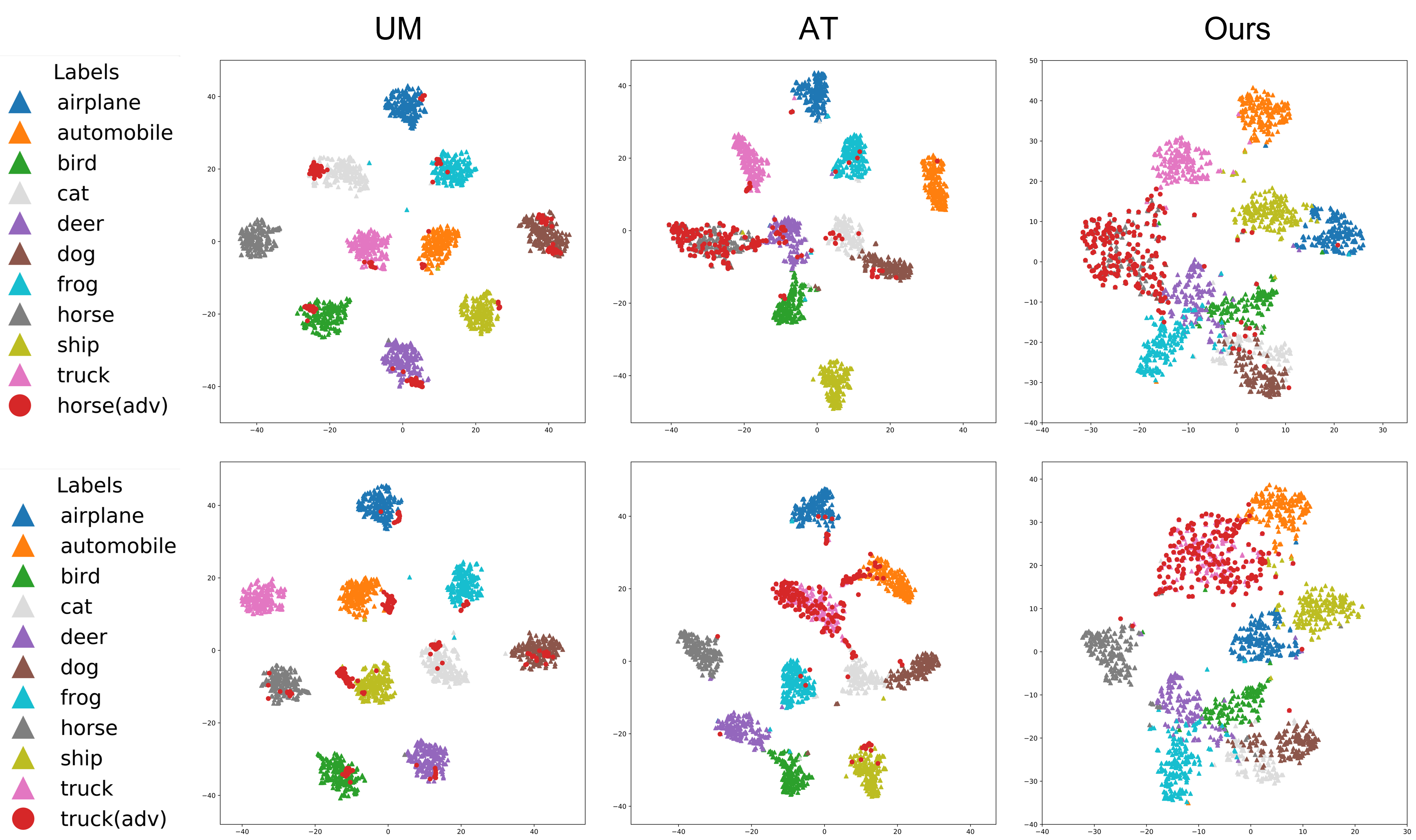}
  \caption{t-SNE plots for illustrating the representation distributions for all the classes in feature space.}
  \label{fig:tsne} 
\end{figure*}

\begin{figure*}
	\centering
	\includegraphics[width=.8\linewidth]{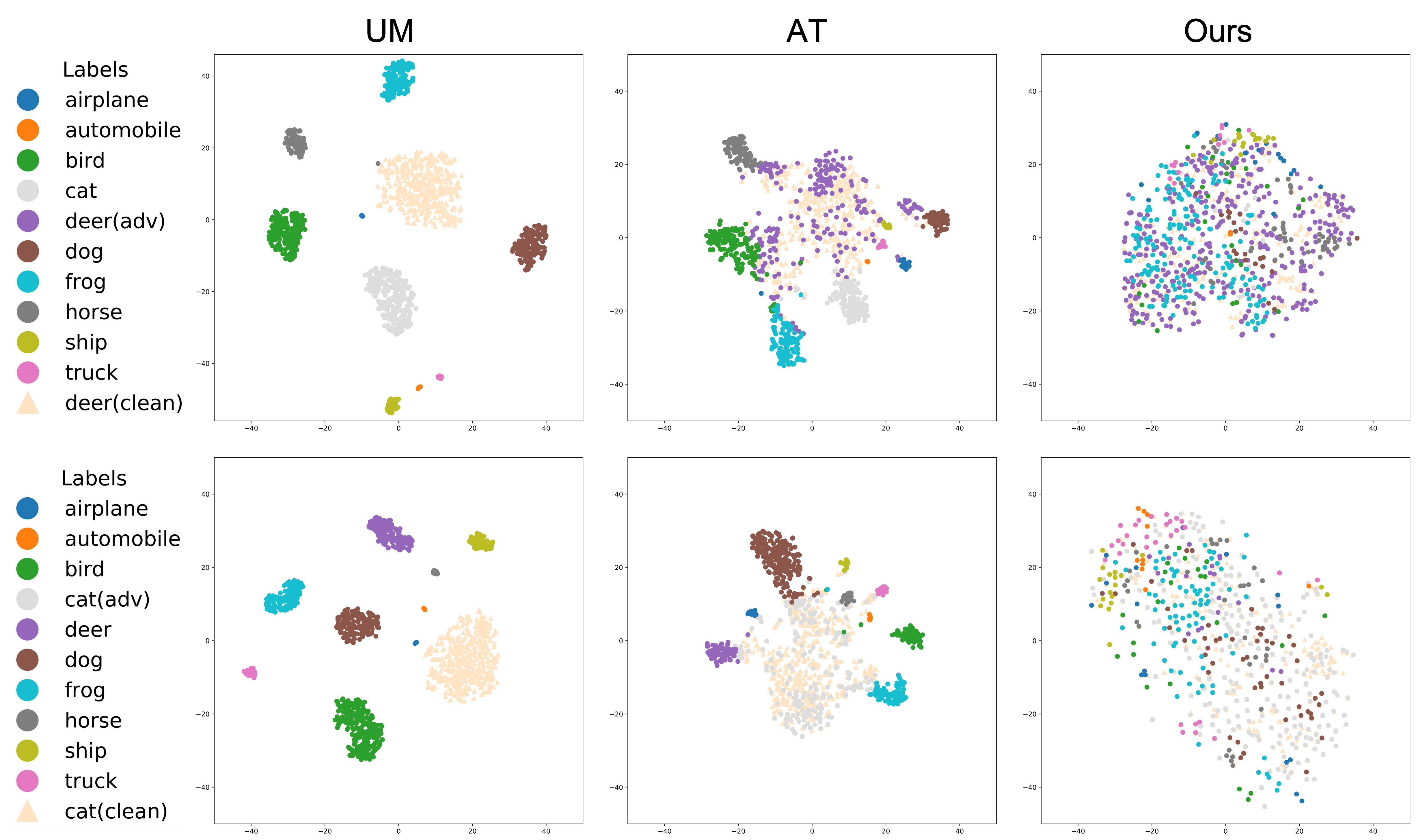} \\
	\caption{t-SNE plots for illustrating the representation distributions of the clean images and adversarial examples for a specific class in feature space.}
	\label{fig:PGD-20-0_4}
\end{figure*}

We visualize the representation distribution of samples in feature space using t-SNE. Specifically, we show the representation distribution for all the classes and for the clean images and adversarial examples in a specific class~\cite{Mao2019}, in Figure~\ref{fig:tsne} and Figure~\ref{fig:PGD-20-0_4}, respectively.

In Figure~\ref{fig:tsne}, the triangle points with different colors represent the clean images in different classes, while the red round points are the AEs from a specific class ( horse and truck) under PGD attack. ``UM", denotes an undefended model, and shows how the adversarial attacks behave on a model that is naturally trained on clean images. PGD attack is able to drop its accuracy to 0, which is also visualized in the first column of Figure~\ref{fig:tsne}, where all the AEs locate far from their original class, and fit into the distributions of other classes. ``AT" shows how adversarial attacks behave on a standard adversarial training model. Many of the adversarial examples are pulled back to their original class but there still are a lot of them mixed with other classes. Our model pulled the most of adversarial examples back to their original class. While we observe that our model shortened the gap between classes for clean images and it may be the reason that the classification performance of our model drops a lot on clean images, we believe it rarely affects the performance on adversarial examples based on the results shown in Table.~\ref{tab:cifar10_result}. For the clean image, as discussed 
previously, our model can keep the same performance as the naturally trained model by using a simple adversarial example detection strategy, thanks to the extraordinary performance of the discriminator and domain specific features obtained from our model.

In Figure~\ref{fig:PGD-20-0_4}, the flesh color triangles are clean images of a specific class and round points are adversarial examples crafted from the clean images of the same class. The colors represent the different predictions of those adversarial examples. In ``UM" has no ability to defend against the attacks and all adversarial examples are falsely predicted. The features of adversarial examples are away from the distribution of natural images and clustered based on the prediction. For ``AT", some of the adversarial examples can be correctly classified. Although the attacks are pulled closer to the location of natural images, the successful attacks are still away and show aggregation based on class. For our model, the natural images and adversarial attacks share almost the same distribution and there is no distinct gathering for different categories, representing that the robust encoder of our model does extract the invariant features between natural images and adversarial examples, that is, the robust features.

\subsection{Discussion}
In~\cite{athalye2018obfuscated}, the authors provided constructive insights about gradient obfuscation in adversarial robustness. Gradient obfuscation is a gradient masking that leads to a false sense of security in defenses against adversarial examples. Based on the analysis in that paper, we conclude that the robustness of our model is not from gradient obfuscation, for the following three reasons: (1) We provide the performance of the model under attacks with different iterations in Figure.~\ref{fig:it}, which shows monotonically declining accuracy along with the attack iterations increasing. (2) In Table~\ref{tab:cifar10_result}, we show that the black-box attacks ($83.04\%$) have a lower success rate (higher accuracy) than white-box attacks ($57.97\%$). (3) We evaluate our model against a gradient-free attack SPSA~\cite{uesato2018adversarial} and obtain an accuracy of 80.21, which is higher than gradient-based attacks ($57.97\%$ for PGD). 

\section{Conclusion}
In this paper, we aimed to explore the robust features, which are not affected by the adversarial perturbations, to improve the model's adversarial robustness as well as generalizability to defend against other unseen attacks. To this end, we proposed a novel three branches feature disentanglement model to segregate the robust features out of non-robust features and domain specific features. The robust features obtained from our model improve the model's adversarial robustness compared to the state-of-the-art approaches through extensive experiments on three widely used datasets, i.e., CIFAR-10, CIFAR-100 and TinyImageNet, under diverse attacks. Moreover, the domain discriminator and domain specific features from our model, as side products, is able to identify the clean images and adversarial examples almost perfectly without additional computational cost. With that, a simple strategy can be applied to specify different classifiers for clean images and adversarial examples, such that our model has almost no sacrifice on the accuracy of clean images.


\bibliographystyle{IEEEtran}
\bibliography{mybib}




 




\vfill

\end{document}